\documentclass[11pt,a4paper]{article}
\usepackage[hyperref]{acl2019}
\usepackage{times}
\usepackage{latexsym}
\usepackage{amsmath}
\usepackage{amsfonts}
\usepackage{url}
\usepackage{subcaption}
\usepackage{graphicx}
\usepackage{booktabs} 

\aclfinalcopy 

\newcommand{\domX}{\mathcal{X}}
\newcommand{\domY}{\mathcal{Y}}
\newcommand\blfootnote[1]{%
  \begingroup
  \renewcommand\thefootnote{}\footnote{#1}%
  \addtocounter{footnote}{-1}%
  \endgroup
}

\title{Microsoft Research Asia's Systems for WMT19}

\author{Yingce Xia, Xu Tan, Fei Tian, Fei Gao, \\
\textbf{Weicong Chen, Yang Fan, Linyuan Gong, Yichong Leng,}\\
\textbf{Renqian Luo, Yiren Wang, Lijun Wu, Jinhua Zhu,} \\ 
\textbf{Tao Qin$^*$, Tie-Yan Liu} \\
Microsoft Research Asia \\
}

\date{}

\begin{document}
\maketitle
\begin{abstract}
We Microsoft Research Asia made submissions to 11 language directions in the WMT19 news translation tasks. We won the first place for 8 of the 11 directions and the second place for the other three. Our basic systems are built on Transformer, back translation and knowledge distillation. We integrate several of our rececent techniques to enhance the baseline systems: multi-agent dual learning (MADL), masked sequence-to-sequence pre-training (MASS), neural architecture optimization (NAO), and soft contextual data augmentation (SCA).\blfootnote{*Corresponding author. This work was conducted at Microsoft Research Asia.} 
\end{abstract}

\section{Introduction}

We participated in the WMT19 shared news translation task in 11 translation directions. We achieved first place for 8 directions: German$\leftrightarrow$English, German$\leftrightarrow$French, Chinese$\leftrightarrow$English, English$\rightarrow$Lithuanian, English$\rightarrow$Finnish, and Russian$\rightarrow$English, and  three other directions were placed second (ranked by teams), which included Lithuanian$\rightarrow$English, Finnish$\rightarrow$English, and English$\rightarrow$Kazakh.

Our basic systems are based on Transformer, back translation and knowledge distillation. We experimented with several techniques we proposed recently. In brief, the innovations we introduced are:
\paragraph{Multi-agent dual learning (MADL)} The core idea of dual learning is to
leverage the duality between the primal task (mapping from domain $\domX$ to domain
$\domY$) and dual task (mapping from domain $\domY$ to $\domX$ ) to boost the performances of
both tasks. MADL~\cite{wang2019multiagent} extends
the dual learning~\cite{he2016dual,dsl} framework by introducing multiple primal and dual models. It was integrated into our submitted systems for German$\leftrightarrow$English and German$\leftrightarrow$French translations.

\paragraph{Masked sequence-to-sequence pretraining (MASS)} Pre-training and fine-tuning have achieved great success
in language understanding. MASS~\citep{song2019mass}, a pre-training method designed for language generation, adopts the encoder-decoder framework
to reconstruct a sentence fragment given the remaining part of the sentence: its encoder takes
a sentence with randomly masked fragment (several consecutive tokens) as input, and its decoder
tries to predict this masked fragment. It was integrated into our submitted systems for Chinese$\to$English and  English$\rightarrow$Lithuanian translations.

\paragraph{Neural architecture optimization (NAO)} As well known, the evolution of neural network architecture plays a key role in advancing neural machine translation. Neural architecture optimization (NAO), our newly proposed method \cite{luo2018neural}, leverages the power of a gradient-based method to conduct optimization and guide the creation of better neural architecture in a continuous and more compact space given the historically observed architectures and their performances. It was applied in English$\leftrightarrow$Finnish translations in our submitted systems. 

\paragraph{Soft contextual data augmentation (SCA)} While data augmentation is an important trick to boost the accuracy of deep learning methods in computer vision tasks, its study in natural  language  tasks  is  relatively limited. SCA  \cite{zhu2019soft} softly augments  a  randomly  chosen  word  in  a  sentence by  its  contextual  mixture  of  multiple  related words, i.e., replacing the one-hot  representation  of  a  word  by  a  distribution provided by a language model over the vocabulary. It was applied in Russian$\rightarrow$English translation in our submitted systems. 


\section{Our Techniques}
\subsection{Multi-agent dual learning (MADL)}
MADL is an enhanced version of dual learning~\citep{he2016dual,wang2018dual}. It leverages $N$ primal translation models $f_i$ and $N$ dual translation models $g_j$ for training, and eventually outputs one $f_0$ and one $g_0$ for inference, where $f_i:\domX\mapsto\domY,g_j:\domY\mapsto\domX$, $i,j\in\{0,1,\cdots,N-1\}$. All these models are pre-trained on bilingual data . The $i$-th primal model $f_i$ has a non-negative weight $\alpha_i$ and the $j$-th dual model $g_i$ has a non-negative weight $\beta_j$. All the $\alpha_\cdot$'s and $\beta_\cdot$'s are hyper-parameters. Let $F_\alpha$ denote a combined translation model from $\domX$ to $\domY$, and $G_\beta$  a combined translation model from $\domY$ to $\domX$, 
\begin{equation}
\begin{aligned}
& F_\alpha=\sum_{i=0}^{N-1}\alpha_if_i,G_\beta=\sum_{j=0}^{N-1}\beta_jg_j;\\
\text{s.t.} & \sum_{i=0}^{N-1}\alpha_i=1;\;\sum_{j=0}^{N-1}\beta_j=1.
\end{aligned}
\end{equation}
$F_\alpha$ and $G_\beta$ work as follows: for any $x\in\domX$ and $y\in\domY$,
\begin{equation*}
\begin{aligned}
F_\alpha(x):&\;\hat{y}=\text{argmax}_{\tilde{y}\in\domY}\sum_{i=0}^{N-1}\alpha_i\log P(\tilde{y}|x;f_i);\\
G_\alpha(y):&\;\hat{x}=\text{argmax}_{\tilde{x}\in\domX}\sum_{j=0}^{N-1}\beta_j\log P(\tilde{x}|y;g_j).
\end{aligned}
\end{equation*}
Let $\mathcal{B}$ denote the bilingual dataset. Let $\mathcal{M}_x$ and $\mathcal{M}_y$ denote the monolingual data of $\domX$ and $\domY$. The training objective function of MADL can be written as follows:
\begin{equation}
\begin{aligned}
\min_{f_0,\,g_0} &-\frac{1}{\vert\mathcal{B}\vert}\sum_{(x,y)\in\mathcal{B}}\log P(y|x;f_0) \\
&-\frac{1}{\vert\mathcal{B}\vert}\sum_{(x,y)\in\mathcal{B}}\log P(x|y;g_0) \\
&-\frac{1}{\vert\mathcal{M}_x\vert}\sum_{x\in\mathcal{M}_x}\log P(x|G_\beta(F_\alpha(x))) \\
  &- \frac{1}{\vert\mathcal{M}_y\vert}\sum_{y\in\mathcal{M}_y}\log P(y|F_\alpha(G_\beta(y))).
\end{aligned}
\label{eq:madl_framework}
\end{equation}
Note that $f_{>0}$ and $g_{>0}$ will not be optimized during training and we eventually output $f_0$ and $g_0$ for translation. More details can be found in~\cite{wang2019multiagent}.

\subsection{Masked sequence-to-sequence pre-training (MASS)}
MASS is a pre-training method for language generation. For machine translation, it can leverage monolingual data in two languages to pre-train a translation model. Given a sentence $x \in \mathcal{X}$, we denote $x^{\setminus u:v}$ as a modified version of $x$ where its fragment from position $u$ to $v$ are masked, $0<u<v<m$ and $m$ is the number of tokens of sentence $x$. We denote $k=v-u+1$ as the number of tokens being masked from position $u$ to $v$. We replace each masked token by a special symbol $[\mathbb{M}]$, and the length of the masked sentence is not changed. $x^{u:v}$ denotes the sentence fragment of $x$ from $u$ to $v$.
	
MASS pre-trains a sequence to sequence model by predicting the sentence fragment $x^{u:v}$ taking the masked sequence $x^{\setminus u:v}$ as input. We use the log likelihood as the objective function:
	\begin{equation}
	\begin{aligned}
	\small
	\label{equ_mass_unsup}
	L(\theta; \mathcal{X}) 
	& = \frac{1}{|\mathcal{X}|}\Sigma_{x \in \mathcal{X}}\log P(x^{u:v}|x^{\setminus u:v};\theta), \\
	L(\theta; \mathcal{Y}) 
	& = \frac{1}{|\mathcal{Y}|}\Sigma_{y \in \mathcal{Y}}\log P(y^{u:v}|y^{\setminus u:v};\theta), \\
	\end{aligned}
	\end{equation}
where $\mathcal{X}$, $\mathcal{Y}$ denote the source and target domain. In addition to zero/low-resource setting~\citep{leng2019unsupervised}, we also extend MASS to supervised setting where bilingual sentence pair $(x, y) \in (\mathcal{X}, \mathcal{Y})$ can be leveraged for pre-training. The log likelihood in the supervised setting is as follows:
\begin{equation}
\begin{aligned}
\small
\label{equ_mass_sup}
&L(\theta; (\mathcal{X}, \mathcal{Y})) = \Sigma_{(x, y) \in (\mathcal{X}, \mathcal{Y})} ( \log P(y|x^{\setminus u:v};\theta) \\
&+\log P(x|y^{\setminus u:v};\theta) \\
&+\log P(x^{u:v}|[x^{\setminus u:v}; y^{\setminus u:v}];\theta) \\
&+\log P(y^{u:v}|[x^{\setminus u:v}; y^{\setminus u:v}];\theta) \\
&+\log P(y^{u:v}|x^{\setminus u:v};\theta) +\log P(x^{u:v}|y^{\setminus u:v};\theta)). \\
\end{aligned}
\end{equation}
where $[\cdot;\cdot]$ represents the concatenation operation. $P(y|x^{\setminus u:v};\theta)$ and $P(x|y^{\setminus u:v};\theta)$ denote the probability of translating a masked sequence to another language, which encourage the encoder to extract meaningful representations of unmasked input tokens in order to predict the masked output sequence. $P(x^{u:v}|[x^{\setminus u:v}; y^{\setminus u:v}];\theta)$ and $P(y^{u:v}|[x^{\setminus u:v}; y^{\setminus u:v}];\theta)$ denote the probability of generating the masked source/target segment given both the masked source and target sequences, which encourage the model to extract cross-lingual information. $P(y^{u:v}|x^{\setminus u:v};\theta)$ and $P(x^{u:v}|y^{\setminus u:v};\theta)$ denote the probability of generating the masked fragment given only the masked sequence in another language. More details about MASS can be found in ~\citet{song2019mass}.


\subsection{Neural architecture optimization (NAO)}
\label{subsec:nao}

\begin{figure*}[t]
\centering
\includegraphics[width=0.90\linewidth]{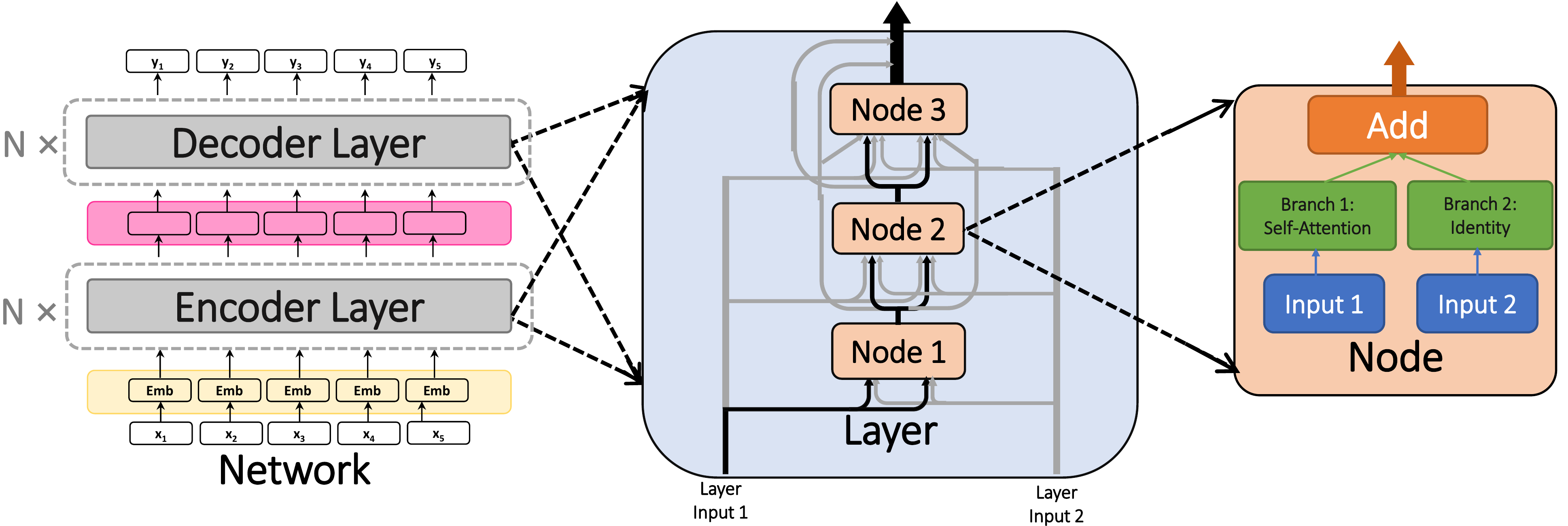}
\caption{Visualization of different levels of the search space, from the \emph{network}, to the \emph{layer}, to the \emph{node}. For each of the different layers, we search its unique \emph{layer} space. The lines in the middle part denote all possible connections between the three nodes (constituting the \emph{layer} space) as specified via each architecture, while among them the deep black lines indicate the particular connection in \emph{Transformer}. The right part similarly contains the two branches used in Node2 of \emph{Transformer}. }
\label{fig:search_space}
\end{figure*}

NAO~\cite{luo2018neural} is a gradient based neural architecture search (NAS) method. It contains three key components: an encoder, an accuracy predictor, and a decoder, and optimizes a network architecture as follows. (1) The encoder maps a network architecture $x$ to an embedding vector $e_x$ in a continuous space $\mathcal{E}$. (2) The predictor, a function $f$, takes $e_x\in \mathcal{E}$ as input and predicts the dev set accuracy of the architecture $x$. We perform a gradient ascent step, i.e., moving $e_x$ along the direction specified via the gradient $\frac{\partial f}{\partial e_x}$, and get a new embedding vector $e_{x'}$:
\begin{equation}
\small
e_{x'}=e_x + \eta\frac{\partial f}{\partial e_x},
\end{equation}where $\eta$ is the step size. (3) The decoder is used to map $e_{x'}$ back to the corresponding architecture $x'$. The new architecture $x'$ is assumed to have better performance compared with the original one $x$ due to the property of gradient ascent. NAO repeats the above three steps, and sequentially generates better and better architectures.


To learn high-quality encoder, decoder and performance prediction function, it is essential to have a large quantity of paired training data in the form of $(x,y)$, where $y$ is the dev set accuracy of the architecture $x$. To reduce computational cost, we share weights among different architectures~\cite{ENAS} to aid the generation of such paired training data.

We use NAO to search powerful neural sequence-to-sequence architectures. The search space is illustrated in Fig.~\ref{fig:search_space}. Specifically, each \emph{network} is composed of $N$ encoder layers and $N$ decoder layers. We set $N=6$ in our experiments. Each encoder \emph{layer} further contains $2$ \emph{nodes} and each decoder layer contains $3$ nodes. The \emph{node} has two branches, respectively taking the output of other node as input, and applies a particular operator (OP), for example, identity, self-attention and convolution, to generate the output. The outputs of the two branches are added together as the output of the \emph{node}. Each encoder layer contains two nodes while each decoder layer has three. For each layer, we search: 1) what is the operator at each branch of every node. For a comprehensive list of different OPs, please refer to the Appendix of this paper; 2) the topology of connection between nodes within each layer. In the middle part of Fig.~\ref{fig:search_space}, we plot possible connections within the nodes of a layer specified by all candidate architectures, with a particular highlight of Transformer~\cite{transformer}.

To construct the final network, we do not adopt the typically used way of stacking the same layer multiple times. Instead we assume that layers in encoder/decoder could have different architectures and directly search such personalized architecture for each layer. We found that such a design significantly improves the performance due to the more flexibility.

\subsection{Soft contextual data augmentation (SCA)}

SCA is a data augmentation technology for NMT \cite{zhu2019soft}, which replaces a randomly chosen word in a sentence with its \textit{soft version}.  For any word $w \in V$, its soft version is a  distribution over the vocabulary of $|V|$ words: $P(w) = (p_1(w), p_2(w), ..., p_{|V|}(w))$, where $p_j(w) \ge 0$ and $\sum_{j=1}^{|V|}p_j(w) = 1$.

Given the distribution $P(w)$, one may simply sample a word from this distribution to replace the original word $w$. Different from this method, we directly use this distribution vector to replace the randomly chosen word $w$ from the original sentence. Suppose $E$ is the embedding matrix of all the $|V|$ words. The embedding of the soft version of $w$ is 
\begin{equation}
    e_{w} = P(w)E = \sum_{j=0}^{\vert V \vert}p_{j}(w) E_j,
\end{equation}
which is the expectation of word embeddings over the distribution.

In our systems, we leverage a pre-trained language model to compute $P(w)$ and condition on all the words preceding $w$. That is, for the $t$-th word $x_t$ in a sentence, we have 
$$p_j(x_t)=LM(v_j|x_{<t}),$$
where $LM(v_j|x_{<t})$ denotes the probability of the $j$-th word $v_j$ in the vocabulary appearing after the sequence $x_1, x_2, \cdots, x_{t-1}$. The language model is pre-trained using the monolingual data. 

\section{Submitted Systems}
\subsection{English$\leftrightarrow$German}\label{sec:en-de}
We submit constrained systems to both English to German and German to English translations, with the same techniques.  
\newcommand{\Mx}{\mathcal{M}_{\text{en}}}
\newcommand{\My}{\mathcal{M}_{\text{de}}}
\paragraph{Dataset} We concatenate ``Europarl v9'', ``News Commentary v14'',  ``Common Crawl corpus'' and ``Document-split Rapid corpus'' as the basic bilingual dataset (denoted as $\mathcal{B}_0$). Since ``Paracrawl'' data is noisy, we select 20M bilingual data from this corpus using the script \texttt{filter\_interactive.py}\footnote{Scripts at \url{https://tinyurl.com/yx9fpoam}}. The two parts of bilingual data are concatenated together (denoted as $\mathcal{B}_1$). We clean $\mathcal{B}_1$ by normalizing the sentences, removing non-printable characters, and tokenization. We share a vocabulary for the two languages and apply BPE for word segmentation with $35000$ merge operations. (We tried different BPE merge operations but found no significant differences.) For monolingual data, we use $120M$ English sentences (denoted as $\Mx$) and $120M$ German sentences (denoted as $\My$) from Newscrawl, and preprocess them in the same way as bilingual data. We use newstest 2016 and the validation set and newstest 2018 as the test set.

\paragraph{Model Configuration} We use the PyTorch implementation of Transformer\footnote{\url{https://github.com/pytorch/fairseq}}. We choose the {\em Transformer\_big} setting, in which both the encoder and decoder are of six layers. The dropout rate is fixed as $0.2$. We set the batchsize as $4096$ and the parameter \texttt{--update-freq} as $16$. We apply Adam~\cite{kingma2015adam} optimizer with learning rate $5\times10^{-4}$.

\paragraph{Training Pipeline} The pipeline consists of three steps: 

1. Pre-train two English$\to$German translation models (denoted as $\bar{f}_1$ and $\bar{f}_2$) and two German$\to$English translation models (denoted as $\bar{g}_1$ and $\bar{g}_2$) on $\mathcal{B}_1$; pre-train another English$\to$German (denoted as $\bar{f}_3$) and German$\to$English (denoted as $\bar{g}_3$) on $\mathcal{B}_0$. 

2. Apply back translation following~\cite{sennrich2016improving,edunov2018understanding}. We back-translate $\Mx$ and $\My$ using $\bar{f}_3$ and $\bar{g}_3$ with beam search, add noise to the translated sentences~\cite{edunov2018understanding}, merge the synthetic data with $\mathcal{B}_1$, and train one English$\to$German model $f_0$ and one German$\to$English model $g_0$ for seven days on eight V100 GPUs. 
   
3. Apply MADL to $f_0$ and $g_0$. That is, the $F_\alpha$ in Eqn.\eqref{eq:madl_framework} is specified as the combination of $f_0,\bar{f}_1,\bar{f}_2$ with equal weights; and $G_\beta$ consists of $g_0,\bar{g}_1,\bar{g}_2$. During training, we will only update $f_0$ and $g_0$. To speed up training, we randomly select $20M$ monolingual English and German sentences from $\Mx$ and $\My$ respectively instead of using all monolingual sentences. The eventual output models are denoted as $f_1$ and $g_1$ respectively. This step takes $3$ days on four P40 GPUs.

\paragraph{Results}
\begin{table}[!htbp]
	\centering
	\caption{Results of English$\leftrightarrow$German by sacreBLEU.}
	\begin{tabular}{ccccc}
		\hline
		& \multicolumn{2}{c}{En$\to$De} & \multicolumn{2}{c}{De$\to$En} \\
		\hline
		& news16 & news18 & news16 & news18 \\
		\hline
		baseline & $37.4$ & $45.6$ & $41.9$ & $44.9$ \\
		BT & $39.2$ & $47.4$ & $45$ & $47.1$ \\
		MADL & $41.9$ & $50.4$ & $47.4$ & $49.1$ \\
		\hline
	\end{tabular}
	\label{tab:wmt:en-de}
\end{table}

The results are summarized in Table~\ref{tab:wmt:en-de}, which are evaluated by sacreBLEU\footnote{\url{https://github.com/mjpost/sacreBLEU}}. The baseline is the average accuracy of models using only bitext, i.e.,  $\bar{f}_1$ and $\bar{f}_2$ for English$\to$German translation and $\bar{g}_1$ and $\bar{g}_2$ for German$\to$English, and BT is the accuracy of the model after back-translation training. As can be seen, back translation improves accuracy. For example, back-translation boosts the BLEU score from $45.6$ to $47.4$ on news18 English$\to$German translation, which is $1.8$ point improvement. MADL further boosts BLEU to $50.4$, obtaining another $3$-point improvement, demonstrating the effectiveness of our method.

For the final submission, we accumulate many translation models (trained using bitext, back translation, and MADL, with different random seeds) and do knowledge distillation on the source sentences from WMT14 to WMT19 test sets. Take English$\to$German translation as an example. Denote the English inputs as $\mathcal{T}=\{s_i\}_{i=1}^{N_T}$, where $N_T$ is the size of the test set. For each $s$ in $\mathcal{T}$, we translate $s$ to $d^\prime$ using $M$ English$\to$German models and eventually obtain
\begin{equation*}
\mathcal{E}=\{(s_i,f^{(j)}(s_i))|s\in\mathcal{T}\}_{i,j},i\in[N_T],j\in[M],
\end{equation*}
where $f^{(j)}$ is the $j$-th translation model we accumulated, $\mathcal{T}$ is the combination of inputs from WMT14 to WMT19. After obtaining $\mathcal{E}$, we randomly select $N_TM$ bitext pairs (denoted as $\mathcal{B}_2$) from $\mathcal{B}_1$ and finetune model $f_1$ on $\mathcal{B}_2\cup\mathcal{E}$. We stop tuning when the BLEU scores of WMT16 (i.e., the validation set) drops. 

We eventually obtain $44.9$ BLEU score for English$\to$German and $42.8$ for German$\to$English on WMT19 test sets and are ranked in the first place in these two translation tasks.

\subsection{German$\leftrightarrow$French}
For German$\leftrightarrow$French translation, we follow a similar process as the one used to  English$\leftrightarrow$German tasks introduced in Section~\ref{sec:en-de}. We merge the ``commoncrawl'', ``europarl-v7'' and part of ``de-fr.bicleaner07'' selected by \texttt{filter\_interactive.py} as the bilingual data. We collect $20M$ monolingual sentences for French and $20M$ for German from newscrawl. The data pre-processing rule and training procedure are the same as that used in Section~\ref{sec:en-de}. We split $9k$ sentences from the ``dev08\_14'' as the validation set and use the remaining ones as the test set.

The results of German$\leftrightarrow$French translation on the test set are summarized in Table~\ref{tab:wmt:de-fr}.

\begin{table}[!htbp]
\centering
\caption{Results of German$\leftrightarrow$French by sacreBLEU.}
\begin{tabular}{ccc}
\hline
& De$\to$Fr & Fr$\to$De\\
\hline
baseline & $29.5$ & $23.4$ \\
MADL & $31.5$ & $24.9$ \\
\hline
\end{tabular}
\label{tab:wmt:de-fr}
\end{table}

Again, our method achieves significant improvement over the baselines. Specifically, MADL boosts the baseline of German$\to$French and French$\to$German by $2$ and $1.5$ points respectively.

Our submitted German$\to$French is a single system trained by MADL, achieving $37.3$ BLEU on WMT19. The French$\to$German is an ensemble of three independently trained models, achieving $35.0$ BLEU score. Our systems are ranked in the first place for both German$\to$French and French$\to$German in the leaderboard.

\subsection{Chinese$\to$English}\label{sec:zh-en}
\paragraph{Dataset}
For Chinese$\to$English translation, we use all the bilingual and monolingual data provided by the WMT official website, and also extra bilingual and monolingual data crawled from the web. We filter the total 24M bilingual pairs from WMT using the script \texttt{filter\_interactive.py} as described in Section~\ref{sec:en-de} and get 18M sentence pairs. We use the Chinese  monolingual data from XMU monolingual corpus\footnote{http://nlp.nju.edu.cn/cwmt-wmt/} and English monolingual data from News Crawl as well as the English sentences from all English-XX language pairs in WMT. We use 100M additional parallel sentences drawn from UN data, Open Subtitles and Web crawled data, which is filtered using the same filter rule described above, as well as fast align and in/out-domain filter. Finally we get 38M bilingual pairs. We also crawled 80M additional Chinese monolingual sentences from Sougou, China News, Xinhua News, Sina News, Ifeng News, and 2M English monolingual sentences from China News and Reuters. We use newstest2017 and newstest2018 on Chinese-English as development datasets.

We normalize the Chinese sentence from SBC case to DBC case, remove non-printable characters and tokenize with both Jieba\footnote{https://github.com/fxsjy/jieba} and PKUSeg\footnote{https://github.com/lancopku/PKUSeg-python} to increase diversity. For English sentences, we remove non-printable characters and tokenize with Moses tokenizer\footnote{https://github.com/moses-smt/mosesdecoder/blob/maste r/scripts/tokenizer/tokenizer.perl}. We follow previous practice~\citep{hassan2018achieving} and apply Byte-Pair Encoding (BPE)~\citep{sennrich2015neural} separately for Chinese and English, each with 40K vocabulary.

\paragraph{MASS Pre-training}
We pre-train MASS (Transfomer\_big) with both monolingual and bilingual data. We use 100M Chinese and 300M English monolingual sentences for the unsupervised setting (Equation~\ref{equ_mass_unsup}), and with a total of 18M and 56M bilingual sentence pairs for the supervised settings (Equation~\ref{equ_mass_sup}). We share the encoder and decoder for all the losses in Equation~\ref{equ_mass_unsup} and~\ref{equ_mass_sup}. We then fine-tune the MASS pre-trained model on both 18M and 56M bilingual sentence pairs to get the baseline translation model for both Chinese$\to$English and English$\to$Chinese.

\paragraph{Back Translation and Knowledge Distillation}
We randomly choose 40M monolingual sentences for Chinese and English respectively for back translation~\citep{sennrich2016improving,he2016dual} and knowledge distillation~\citep{SeqKD,tan2018multilingual}. We iterate back translation and knowledge distillation multiple times, to gradually boost the performance of the model.

\paragraph{Results} The results on newstest2017 and newstest2018 are shown in Table~\ref{table_zhen}. We list two baseline Transformer\_big systems which use 18M bilingual data (constraint) and 56M bilingual data (unconstraint) respectively. The pre-trained model achieves about 1 BLEU point improvement after fine-tuning on both 18M and 56M bilingual data. After iterative back translation (BT) and knowledge distillation (KD), as well as re-ranking, our system achieves 30.8 and 30.9 BLEU points on newstest2017 and newstest2018 respectively.
\begin{table}[htbp]
\centering
\begin{tabular}{l   c  c}
\toprule
System            & newstest17 & newstest18 \\ 
\midrule
Baseline (18M)    & 24.2 & 24.5 \\
+ MASS (18M) & 25.2 & 25.4 \\
Baseline (56M)    & 26.9 & 27.0 \\
+ MASS (56M)        & 28.0 & 27.8 \\
+ Iterative BT/KD & 30.4 & 30.5 \\ 
+ Reranking       & 30.8 & 30.9 \\
\bottomrule
\end{tabular}
\caption{BLEU scores on  Chinese$\to$English test sets.}
\label{table_zhen}
\end{table}

\paragraph{WMT19 Submission}
For the WMT19 submission, we conduct fine-tuning and speculation to further boost the accuracy by using the source sentences in the WMT19 test set. We first filter the bilingual as well as pseudo-generated data according to the relevance to the source sentences. We use the filter method in~\citet{deng2018alibaba} and continue to train the model on the filtered data. Second, we conduct speculation on the test source sentences following the practice in~\citet{deng2018alibaba}. The final BLEU score of our submission is 39.3, ranked in the first place in the leaderboard.

\subsection{English$\leftrightarrow$Lithuanian}
For English$\leftrightarrow$Lithuanian translation, we follow the similar process as that for Chinese$\to$English task introduced in Section~\ref{sec:zh-en}. We use all the WMT bilingual data, which is 2.24M after filtration. We use the same English monolingual data as used in Chinese-English. We select 100M Lithuanian monolingual data from official commoncrawl and use all the wiki and news Lithuanian monolingual data provided by WMT. In addition, we crawl 5M Lithuanian news data from LRT website\footnote{https://www.lrt.lt/}. We share the BPE vocabulary between English and Lithuanian, and the vocabulary size is 65K.

All the bilingual and monolingual data are used for MASS pre-training, and all the bilingual data are used for fine-tuning. For iterative back translation and knowledge distillation, we split 24M English monolingual data as well as 12M Lithuanian monolingual data into 5 parts through sampling with replacement, to get different models independently so as to increase diversity in re-ranking/ensemble. Each model uses 8M English monolingual data and 6M Lithuanian monolingual data. For our WMT19 submission, different from zh-en, speculation technology is not used.

The BLEU scores on newsdev19 are shown in Table~\ref{table_lten}. Our final submissions for WMT19 achieves 20.1 BLEU points for English$\to$Lithuanian translation (ranked in the first place) and 35.6 for Lithuanian$\to$English translation (ranked in the second place).

\begin{table}[htbp]
\centering
\begin{tabular}{l   c  c}
\toprule
System            & En$\to$Lt & Lt$\to$En \\ 
\midrule
Baseline          & 20.7 & 28.2 \\
MASS + Fine-tune            & 21.5 & 28.7 \\
+ Iterative BT/KD & 28.3 & 33.6 \\ 
+ Reranking       & 29.1 & 34.2 \\
\bottomrule
\end{tabular} 
\caption{BLEU scores for English$\leftrightarrow$Lithuanian on the newsdev19 set.}
\label{table_lten}
\end{table}

\subsection{English$\leftrightarrow$Finnish}
\paragraph{Preprocess} We use the official English-Finnish data from WMT19, including both bilingual data and monolingual data. After de-duplicating, the bilingual data contains $8.8M$ aligned sentence pairs. We share the vocabulary for English and Finnish with $46k$ BPE units. We use the WMT17 and WMT18 English-Finnish test sets as two development datasets, and tune hyper-parameters based on the concatenation of them.

\paragraph{Architecture search} We use NAO to search sequence-to-sequence architectures for English-Finnish translation tasks, as introduced in subsection~\ref{subsec:nao}. We use PyTorch for our implementations.  Due to time limitations, we are not targeting at finding better neural architectures than Transformer; instead we target at models with comparable performance to Transformer, while providing diversity in the reranking process. The whole search process takes $2.5$ days on $16$ P40 GPU cards and the discovered neural architecture, named as NAONet, is visualized in the Appendix.

\paragraph{Train single models} 
\label{subsubsec:single_model}
The final system for English-Finnish is obtained through reranking of three strong model checkpoints, respectively from the Transformer model decoding from left to right (L2R Transformer), the Transformer model decoding from right to left (R2L Transformer) and NAONet decoding from left to right. All the models have 6-6 layers in encoder/decoder, and are obtained using the same process which is detailed as below. 

\emph{Step 1: Base models.} Train two models $P_1(x|y)$ and $P_1(y|x)$ based on all the bilingual dataset ($8.8$M), respectively for English$\to$Finnish and Finnish$\to$English translations.

\emph{Step 2: Back translation.} Do the normal back translation~\cite{sennrich2016improving,he2016dual} using $P_1$ and $P_2$. Specifically we choose $10M$ monolingual English corpus, use $P_1(y|x)$ to generate the $10M$ pseudo bitext with beam search (beam size is set to $5$), and mix it with the bilingual data to continue the training of $P_1(x|y)$. The ratio of mixing is set as $1:1$ through up-sampling. The model obtained through such a process is denoted as $P_2(x|y)$. The same process is applied to the opposite direction and the new model $P_2(y|x)$ is attained. 

\emph{Step 3: Back translation + knowledge distillation.} In this step we generate more pseudo bitext by sequence level knowledge distillation~\cite{SeqKD} apart from using back translation. To be more concrete, as the first step, similar to Step 2, we choose $15M$ monolingual English and Finnish corpus, and generate the translations using $P_2(y|x)$ and $P_2(x|y)$, respectively. The resulting pseudo bitext is respectively denoted as $D_{x\rightarrow y}$ and $D_{y\rightarrow x}$. Then we concatenate all the bilingual data, $D_{x\rightarrow y}$ and $D_{y\rightarrow x}$, and use the whole corpus to train a new English-Finnish model \emph{from scratch}. The attained model is denoted as $P_3(y|x)$. 

\emph{Step 4: Finetune.} In this step we try a very simple data selection method to handle the domain mismatch problem in WMT. We remove all the bilingual corpus from Paracrawl which is generally assumed to be quite noisy~\cite{junczys2016microsoft} and use the remaining bilingual corpus ($4.5M$) to finetune $P_3(y|x)$ for one epoch. The resulting model is denoted as $P_4(y|x)$ which is set as the final model checkpoint. 

\begin{table}[htbp] \centering
\begin{tabular}{ccc}
\hline
          & newstest17 & newstest18 \\ \hline
Baseline      & 26.09 & 16.07 \\
+BT       & 28.84 & 18.54 \\
+BT \& KD   & 29.76 & 19.13 \\
+Finetune & 30.19 & 19.46 \\ \hline
\end{tabular}
\caption{BLEU scores of L2R Transformer on English$\to$Finnish test sets.}
\label{tbl:single_steps}
\end{table}

\begin{table}[htbp]
\begin{tabular}{ccc}
\hline
                & newstest17 & newstest18 \\ \hline
L2R Transformer & 30.19 & 19.46 \\
R2L Transformer & 30.40 & 19.73 \\
NAONet          & 30.54 & 19.58 \\ \hline
\end{tabular}
\caption{The final BLEU scores on English$\to$Finnish test sets, for the three models: L2R Transformer, R2L Transformer and NAONet, after the four steps of training.}
\label{tbl:final_scores}
\end{table}

To investigate the effects of the four steps, we record the resulting BLEU scores on WMT17 and WMT18 test sets in Table~\ref{tbl:single_steps}, taking the L2R Transformer model as an example. Furthermore, we report the final BLEU scores  of the three models after the four steps in Table~\ref{tbl:final_scores}. All the results are obtained via beam size $5$ and length penalty $1.0$. The similar results for Finnish-English translation are shown in Table~\ref{tbl:fe_final_scores}.

\begin{table}[htbp]
\begin{tabular}{ccc}
\hline
                & newstest17 & newstest18 \\ \hline
L2R Transformer & 35.66 & 25.56 \\
R2L Transformer & 35.31 & 25.56 \\
NAONet          & 36.18 & 26.38 \\ \hline
\end{tabular}
\caption{The final BLEU scores on Finnish$\to$English test sets, for the three models: L2R Transformer, R2L Transformer and NAONet, after the four steps of training.}
\label{tbl:fe_final_scores}
\end{table}

\paragraph{Re-ranking}
We use n-best re-ranking to deliver the final translation results using the three model checkpoints introduced in the last subsection. The beam size is set as $12$. The weights of the three models, as well as the length penalty in generation, are tuned on the WMT-18 test sets. The results are shown in the second row of Table~\ref{tbl:reranking}. 

\begin{table}[]
\begin{tabular}{cccc}
\hline
                                                                 & news17 & news18 & \multicolumn{1}{l}{news19} \\ \hline
\begin{tabular}[c]{@{}c@{}}Re-ranking\\  w/ NAONet\end{tabular}  & 31.48 & 21.21 & 27.4                      \\ \hline
\begin{tabular}[c]{@{}c@{}}Re-ranking \\ w/o NAONet\end{tabular} & 30.82 & 20.79 & /                         \\ \hline
\end{tabular}
\caption{English$\to$Finnish BLEU scores of re-ranking using the three models. ``news'' is short for ``newstest''.}
\label{tbl:reranking}
\end{table}

\begin{table}[]
\begin{tabular}{cccc}
\hline
                                                                 & news17 & news18 & \multicolumn{1}{l}{news19} \\ \hline
\begin{tabular}[c]{@{}c@{}}Re-ranking\\  w/ NAONet\end{tabular}  & 37.54 & 27.51 & 31.9                      \\ \hline
\begin{tabular}[c]{@{}c@{}}Re-ranking \\ w/o NAONet\end{tabular} & 36.83 & 26.99 & /                         \\ \hline
\end{tabular}
\caption{Finnish$\to$English BLEU scores of re-ranking using the three models. }
\label{tbl:fe_reranking}
\end{table}

We would also like to investigate what is the influence of the NAONet to the re-ranking results. To achieve that, in re-ranking we replace NAONet with another model from L2R Transformer, trained with the same process in subsection~\ref{subsubsec:single_model} with the difference only in random seeds, while maintain the other two models unchanged. The results are illustrated in the last row of Table~\ref{tbl:reranking}. From the comparison of the two rows in Table~\ref{tbl:reranking}, we can see the new architecture NAONet discovered via NAO brings more diversity in the ranking, thus leading to better results. We also report the similar results for Finnish-English tasks in Table~\ref{tbl:fe_reranking}.

Our systems achieve $27.4$ for and $31.9$ for English$\to$Finnish and Finnish$\to$English, ranked in the first place and second place (by teams), respectively. 
\subsection{Russian$\to$English}

\paragraph{Dataset} We use the bitext data from the several corpora: ParaCrawl, Common Crawl, News Commentary, Yandex Corpus, and UN Parallel Corpus. We also use News Crawl corpora as monolingual data. The data is filtered by rules such as sentence length, language identification, resulting a training dataset with 16M bilingual pairs and 40M monolingual sentences (20M for English and 20M for Russian). We use WMT17 and WMT18 test set as development data. The two languages use separate vocabularies, each with 50K BPE merge operations. 

\paragraph{Our system} Our final system for Russian$\rightarrow$English translation is a combination of Transformer network \cite{transformer}, back translation \cite{sennrich2016improving}, knowledge distillation \cite{SeqKD}, soft contextual data augmentation \cite{zhu2019soft}, and model ensemble. We use Transformer\_big as network architecture. We first train two models, English$\rightarrow$Russian and Russian$\rightarrow$English respectively, on bilingual pairs as baseline model. Based on these two models, we perform  back translation and knowledge distillation on monolingual data, generating 40M  synthetic data. 
Combining both bilingual and synthetic data, we get a large train corpus with 56M pairs in total. We upsample the bilingual pairs and shuffle the combined corpus to ensure the balance between bilingual and synthetic data. Finally, we train the Russian$\rightarrow$English model from scratch. During the training, we also use soft contextual data augmentation to further enhance training. Following the above procedures, 5 different models are trained and ensembled for final submission. 

\paragraph{Results} Our final submission achieves 40.1 BLEU score, ranked first in the leaderboard. Table \ref{tbl:ru} reports the results of our system on the development set. 

\begin{table}[htbp]\centering
\begin{tabular}{ccc}
\hline
          & newstest17 & newstest18 \\ \hline
Baseline      & 36.5 & 32.6 \\
+BT \& KD   & 40.9 & 35.2 \\
+SCA & 41.7 & 35.6 \\ \hline
\end{tabular}
\caption{Russian$\rightarrow$English BLEU scores. }
\label{tbl:ru}
\end{table}

\subsection{English$\to$Kazakh}
\paragraph{Dataset} We notice that most of the parallel data are out of domain. Therefore, we crawl some external data:

(1) We crawl all news articles from \url{inform.kz}, a Kazakh-English news website. Then we match an English new article to a Kazakh one by matching their images with image hashing. In this way, we find 10K pairs of bilingual news articles. We use their title as additional parallel data. These data are in-domain and useful in training.
    
    (2) We crawl 140K parallel sentence pairs from \url{glosbe.com}. Although most of these sentences are out-of-domain, they significantly extended the size of our parallel dataset and lead to better results.

Because most of our parallel training data are noisy, we filter these data with some rules:
(1) For the \textit{KazakhTV} dataset, we remove any sentence pair with an alignment score less than 0.05. (2) For the \textit{Wiki Titles} dataset, we remove any sentence pair that starts with \textit{User} or \textit{NGC}.
(3) For all datasets, we remove any sentence pair in which the English sentence contains no lowercase alphabets.
   (4) For all datasets, we remove any sentence pair where the length ratio is greater than 2.5:1.

We tokenize all our data using the \href{https://github.com/moses-smt/mosesdecoder}{Moses Decoder}. We learn a shared BPE ~\citep{sennrich2015neural} from all our data (including all WMT19 parallel data, WMT19 monolingual data\footnote{When we learn BPE, English monolingual data is down-sampled to make the number of English sentences roughly the same as the number of Kazakh sentences.}, glosbe, inform.kz news titles, and inform.kz news contents) and get a shared vocabulary of 49,152 tokens. Finally, our dataset consists of 300K bilingual sentence pairs, 700K Kazakh monolingual sentences, and many English monolingual sentences.

\paragraph{Our system} Our model is based on the Transformer~\cite{transformer}. We vary the hyper-parameters to increase the diversity of our model. Our models usually have 6 encoder layers, 6/7 decoder layers, ReLU/GELU~\cite{DBLP:journals/corr/HendrycksG16} activation function, and an embedding dimension of 640.

We train 4 English-Kazakh models and 4 Kazakh-English models with different random seeds and hyper-parameters. Then we apply back-translation~\cite{edunov2018understanding} and knowledge distillation~\citep{SeqKD} for 6 rounds. In each round, we

    1. Sample 4M sentences from English monolingual data and back-translate them to Kazakh with the best EN-KK model (on the dev set) in the previous round.
    
    2. Back-translate all Kazakh monolingual data to English with the best KK-EN model in the previous round.
    
    3. Sample 200K sentences from English monolingual data and translate them to Kazakh using the ensemble of all EN-KK models in the previous round.
    
    4. Train 4 English-Kazakh models with BT data from step 2 and KD data from step 3. We up-sample bilingual sentence pairs by 2x.
    
    5. Train 4 Kazakh-English models with BT data from step 1. We up-sample bilingual sentence pairs by 3x.

\paragraph{Result} Our final submission achieves 10.6 BLEU score, ranked second by teams in the leaderboard.

\section{Conclusions}
This paper describes Microsoft Research Asia's neural machine translation systems for the WMT19 shared news translation tasks. Our systems are built on   Transformer,  back translation  and  knowledge  distillation, enhanced with our recently proposed  techniques:    multi-agent dual learning (MADL), masked sequence-to-sequence  pre-training  (MASS),  neural  architecture optimization (NAO), and soft contextual data augmentation (SCA). Due to time and GPU limitations, we only apply each technique to a subset of translation tasks. We believe combining them together will further improve the translation accuracy and will conduct experiments in the future. Furthermore, some other techniques such as deliberation learning~\citep{xia2017deliberation}, adversarial learning~\citep{wu2018adversarial}, and reinforcement learning~\citep{he2017decoding,wu2018study} could also hep and are worthy of exploration.

\section*{Acknowledgments}
This work is supported by Microsoft Machine Translation team.\\

\bibliography{acl2019}
\bibliographystyle{acl_natbib}



\end{document}